\documentclass{article}
\usepackage{spconf,amsmath,graphicx}
\usepackage{indentfirst}
\usepackage{multirow}
\usepackage{booktabs}
 \usepackage{amssymb}
 \usepackage{color}

 \pdfoutput=1
\title{Semi-UFormer: Semi-supervised Uncertainty-aware Transformer for Image Dehazing}
\name{Ming Tong$^1$, Yongzhen Wang$^1$, Peng Cui$^2$, Xuefeng Yan$^1$, Mingqiang Wei$^1$}
\address{$^1$Nanjing University of Aeronautics and Astronautics\\
$^2$PLA Dalian Naval Academy}
\begin{document}
%
\maketitle
\begin{abstract}
Image dehazing is fundamental yet not well-solved in computer vision. Most cutting-edge models are trained in synthetic data, leading to the poor performance on real-world hazy scenarios.
Besides, they commonly give deterministic dehazed images while neglecting to mine their uncertainty.
To bridge the domain gap and enhance the dehazing performance, we propose a novel semi-supervised uncertainty-aware transformer network, called Semi-UFormer. Semi-UFormer can well leverage both the real-world hazy images and their uncertainty guidance information.
Specifically, Semi-UFormer builds itself on the knowledge distillation framework. Such teacher-student networks effectively absorb real-world haze information for quality dehazing. Furthermore, an uncertainty estimation block is introduced into the model to estimate the pixel uncertainty representations, which is then used as a guidance signal to help the student network produce haze-free images more accurately. Extensive experiments demonstrate that Semi-UFormer generalizes well from synthetic to real-world images.
\end{abstract}
\section{Introduction}
\label{sec:intro}

Images captured in hazy weather often suffer from noticeable visibility degradation, color distortions, and contrast reduction, which further drop the performance of downstream vision-based systems such as object detection, autonomous driving, and traffic surveillance \cite{chen2018domain,choi2017sharpness}. Therefore, restoring clean images from their hazy versions is quite important. 

Existing dehazing efforts can be roughly classified into prior-based \cite{he2010single,berman2016non} and learning-based approaches \cite{li2017aod,chen2019gated}. 
Traditional prior-based studies often exploit hand-crafted image priors to solve the image dehazing problem based on the atmospheric scattering model \cite{narasimhan2002vision}, such as dark channel prior (DCP) \cite{he2010single}, non-local color prior \cite{berman2016non}, etc. Although these algorithms can improve the overall visibility of the image, they are not always reliable due to over-reliance on assumptions.

Recent advances in deep learning open up huge opportunities for image dehazing tasks, and a large number of learning-based approaches have sprung up \cite{li2017aod,chen2019gated,qin2020ffa,song2022vision}. While these algorithms are efficient and can produce promising results on various popular benchmarks, most of them are trained on synthetic data, so they cannot generalize well to real-world scenes due to the existence of domain gap. Recently, several semi-supervised \cite{li2019semi,chen2021psd} and unsupervised \cite{li2021you,wang2022cycle} methods have attempted to solve the domain shift issue via training models on real-world images. Although these approaches alleviate the domain shift issue to some extent, their dehazing capacity is usually limited because the ground-truth of real-world hazy images cannot be used as the reconstruction loss to constrain the training of the network. 

In addition, most learning-based dehazing efforts only produce the final dehazed images without discussing the uncertainty of the results, which is important for recovering edge and texture regions in hazy images. Furthermore, current dehazing models typically treat all pixels equally, but pixels in edge and texture regions obviously contain more visual information than pixels in smooth regions \cite{ning2021uncertainty}, and such pixels tend to have a high degree of uncertainty. Therefore, accurate estimation and appropriate use of image uncertainty can guide the network to focus on pixels with large uncertainty, so that pixels in specific regions can be enhanced to improve the quality of the final dehazed images. 

To resolve these issues, we propose a novel semi-supervised uncertainty-aware transformer network (Semi-UFormer) for image dehazing. Semi-UFormer builds itself on the knowledge distillation framework and benefits from uncertainty guidance information, thus producing much clearer images with well-preserved details. In summary, our contributions are three-fold: (1) A novel semi-supervised uncertainty-aware transformer network called Semi-UFormer is proposed for image dehazing, which leverages both real-world data and uncertainty guidance information to boost the model's dehazing ability. (2) An uncertainty estimation block is exploited to predict the epistemic uncertainty of the dehazed images, which is then used to guide the network to better reconstruct the image texture and edge regions. (3) We leverage knowledge distillation technology to align the feature distributions between synthetic and real data, which can help the network generalize well in real-world scenarios. 

\section{Semi-UFormer}
\label{sec:pagestyle}

\begin{figure}[htb]

\begin{minipage}[b]{1.0\linewidth}
  \centering
  \centerline{\includegraphics[width=8.5cm]{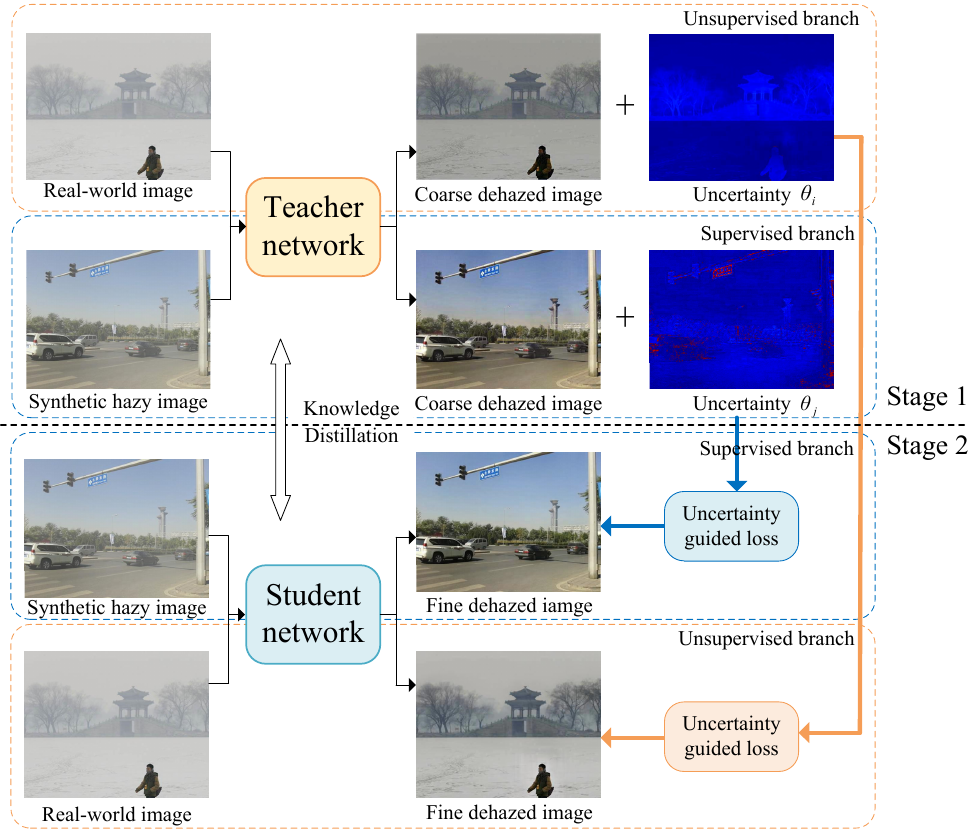}}
\end{minipage}
\caption{Overall architecture of Semi-supervised Uncertainty-aware Transformer Network (Semi-UFormer).}
\label{fig:1}
\end{figure}

Beyond existing image dehazing wisdom, Semi-UFormer fully explores knowledge distillation technology and uncertainty guidance information to help the network produce much clearer images with more confidence. Fig. \ref{fig:1} exhibited the overview of our Semi-UFormer, where the teacher and student network share the same architecture. We first train the teacher network on both synthetic and real-world data to produce the coarse dehazed results and pixel uncertainty map, $\theta$. Then, the student network produces fine dehazed results with the help of uncertainty-guided information $\theta$ and knowledge distillation strategies. The specific dehazing process is described in the following. In the following, we will detail the individual network modules in our Semi-UFormer.


\subsection{Teacher-student Network with Knowledge Distillation}
\label{ssec:subhead}
Unlike existing image dehazing networks that are exclusively trained on synthetic data, we serve the teacher-student network as the semi-supervised framework and utilize knowledge distillation to migrate supervised dehazing knowledge to unsupervised dehazing. 

{\bf Teacher-student network.} The training phase of Semi-UFormer can be divided into two stages, as depicted in Fig. \ref{fig:1}. In stage 1, the teacher network is trained on both synthetic and real data to estimate uncertainty information and coarse dehazed images, where the supervised branch plays a leading role. In stage 2, we first leverage the weights of teacher network to initialize the student model. Then, with the help of the teacher network, the student model retrains on both data and exploits knowledge distillation for better generalization in real-world scenes, where the unsupervised branch plays a leading role. At the same time, the uncertainty $\theta$ supplied by the teacher model is used as additional guidance information to teach the student how to produce fine dehazed images, while the network for estimating uncertainty $\theta$ is frozen.

{\bf Knowledge Distillation.} Considering similar images tend to demonstrate correlation at the high-dimensional feature level, we leverage the teacher-student framework to extract the high-dimensional features of synthetic and real haze. Then, we minimize the KL divergence between these two features to reduce the gap between synthetic and real-world data \cite{cui2022semi}, to help the student model apply the supervised dehazing knowledge to unsupervised haze removal. 


\subsection{Transformer-based Dehazing Network}
\label{ssec:subhead}

Typically, real-world images follow a definite rule and reflect global properties such as contrast ratio and sparsity of dark channel. To capture global information and perform accurate dehazing, we introduced the Dehazeformer block into our network because of the excellent dehazing abilities of the Dehazeformer-Net \cite{song2022vision}. Additionally, to reduce computational costs, the detailed parameters of the Dehazeformer block in this paper are referred to the Dehazeformer-Small in \cite{song2022vision}.

\begin{figure}[htb]

\begin{minipage}[b]{1.0\linewidth}
  \centering
  \centerline{\includegraphics[width=8.5cm]{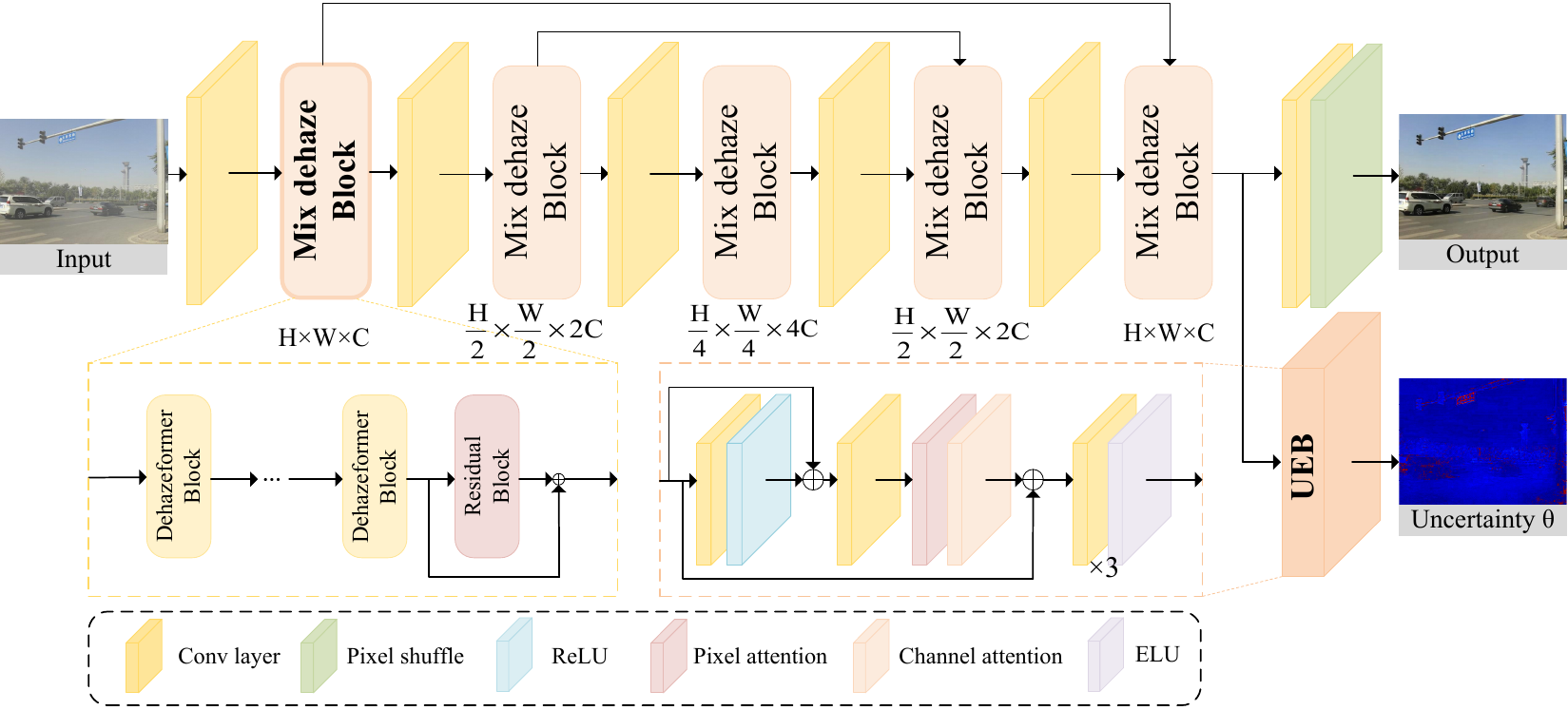}}
\end{minipage}
\caption{Overview of our Transformer-based dehazing network.}
\label{fig:2}
\end{figure}

As exhibited in Fig. \ref{fig:2}, the Transformer-based dehazing network is an enhanced 5-stage U-Net, which consists of three modules: a shallow feature extraction, a Mix DehazeFomer Block, and a reconstruction module.

Shallow feature extraction. A ${3\times3}$ convolutional layer $C_3 (.)$ is first applied to extract shallow feature information from the hazy image $I\epsilon R^{H\times W\times 3}$:
\begin{equation}
\begin{aligned}
F_{shallow}=C_3(I)
\end{aligned}
\end{equation}

Mix DehazeFormer Block (MDB). Next, feature $F_{shallow}$ will be sent to Mix DehazeFormer Block for extracting image global features. In MDB, we first leverage several Dehazeformer blocks $DF(.)$ (see Fig. 3) to extract the global information and then use the residual block without normalization layer $RB(.)$ to fuse the global information, such a hybrid structure is more efficient than using only Transformer blocks. The global feature $F_{global}\epsilon R^{H\times W\times 3} $ is:
\begin{equation}
\begin{aligned}
F_{global}=RB(DF(F_{shallow} )_n )+ DF(F_{shallow} )_n
\end{aligned}
\end{equation}
where $n$ denotes the number of Dehazeformer blocks used in MDB. 



\begin{figure}[htb]

\begin{minipage}[b]{1.0\linewidth}
  \centering
  \centerline{\includegraphics[width=8.5cm]{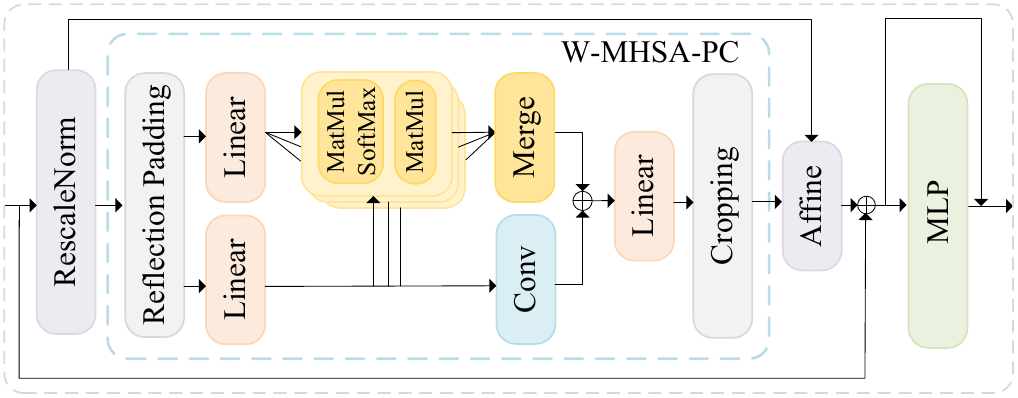}}
\end{minipage}
\caption{Architecture of Dehazeformer block. W-MHSA-PC denotes the window-shifted multi-head self-attention with parallel convolution.}
\label{fig:3}
\end{figure}

Reconstruction module. Finally, a $3\times3$ convolutional layer $C_3 (.)$ and a pixelshuffle layer $P(.)$ are used to produce the haze-free image $J\epsilon R^{H\times W\times 3}$ from the extracted $ F_{global} $:
\begin{equation}
\begin{aligned}
J=P(C_3 (F_{global}))
\end{aligned}
\end{equation}

\subsection{Uncertain Estimated Block and Uncertain Loss}
\label{ssec:subhead}
Theoretically, there are two types of uncertainty in Bayesian modeling: aleatoric uncertainty from the data and epistemic uncertainty from the model. The former is very common in dehazing models, but existing methods ignore exploring it. Therefore, we exploit an uncertainty estimation block (UEB) to predict the uncertainty of dehazing results, which enables the model to focus on regions with rich visual information (e.g., edge areas) to improve the final restoration results.

The prediction process for uncertainty $ \theta $ \cite{ning2021uncertainty} can be expressed as:
\begin{equation}
\begin{aligned}
\hat{J_i}=G_1 (I_i)+\epsilon \theta_i
\end{aligned}
\end{equation}
where $\hat{J_i}$, $G_1 (I_i)$, $\epsilon$ and $\theta_i$ denote the ground-truth, coarse dehazed image, Laplace distribution, and aleatoric uncertainty from the synthetic dehazed image, respectively. For a more accurate prediction of $\theta_i$, we introduce Jeffrey’s prior \cite{figueiredo2001adaptive} into the uncertainty estimation process. For $\hat{J_i}$, $G_1 (I_i)$, the Laplace distribution-characterized log-likelihood function and uncertainty estimation loss $ L_{ue} $ can be expressed as:
\begin{equation}
\begin{aligned}
\ln p(\hat{J_i}| I_i)=-\frac{\left \| \hat{J_i}-G_1(I_i)\right \|_1}{\theta _i}-2\ln \theta _i-\ln 2
\end{aligned}
\end{equation}
\begin{equation}
\begin{aligned}
L_{ue}=\frac{1}{N}\sum_{i=1}^{N}exp(-\ln \theta _i)\left \| \hat{J_i}-I_i\right \|_1+2\ln \theta _i
\end{aligned}
\end{equation}

We employ $L_{ue}$ to estimate the $\theta$ more accurately. Then, through the guidance of $\theta$, we apply the uncertainty-guided loss $L_{ugs}$ to push the network to concentrate more on the reconstruction error area with large uncertainty in the dehazed image, to obtain accurate and confident dehazed results. The formula is shown in (\ref{con:2}). In addition, inspired by the identity loss \cite{zhu2017unpaired}, we incorporate the uncertainty-guided loss $L_{ugu}$ into the unsupervised branch, as shown in (\ref{con:3}).
\begin{equation}
\begin{aligned}
L_{ugs}=\frac{1}{N}\sum_{i=1}^{N}(\ln \theta _i-\min (\ln \theta _i))\left \| \hat{J_i}-G_2(I_i)\right \|_1
\label{con:2}
\end{aligned}
\end{equation}

\begin{equation}
\begin{aligned}
L_{ugu}=\frac{1}{N}\sum_{j=1}^{N}(\ln \theta _j-\min (\ln \theta _j))\left \| J_j-G_2(J_j)\right \|_1
\label{con:3}
\end{aligned}
\end{equation}
where $G_2 (.)$, $J_j$, $\theta_j$ represents the student network, real-world images, and uncertainty from real-world dehazed images, respectively.



\begin{figure*}[htbp]
\centering
\includegraphics[scale=1.2]{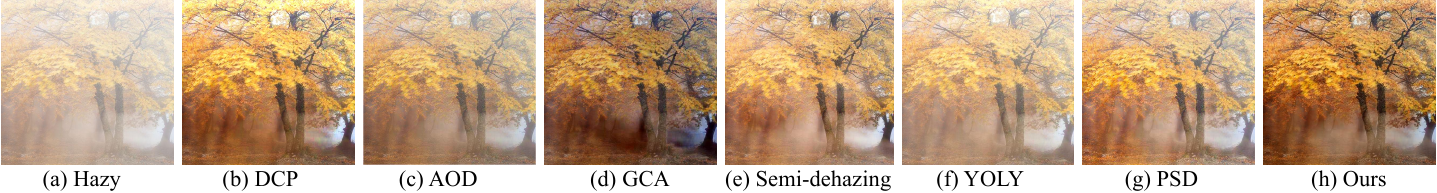}
\caption{Dehazing results on a real-world image.}
\label{fig:4}
\end{figure*}

\subsection{Loss Functions}
\label{sec:loss}

{\bf Teacher network.} The overall loss functions for the teacher network are formulated as:
 \begin{equation}
\begin{aligned}
 L_t = L_{ts} + L_{tu}
\label{con:4}
\end{aligned}
\end{equation}
where $L_{ts}$, $L_{tu}$ refers to the loss functions of the supervised and unsupervised branches, respectively.
 \begin{equation}
\begin{aligned}
L_{ts} = \lambda _1*L_{ue} +\lambda _2*L_{a}
\end{aligned}
\end{equation}
 \begin{equation}
\begin{aligned}
L_{tu} = \lambda _3*L_{ide} +\lambda _4*L_{dc} + \lambda _5*L_{tv}
\label{con:6}
\end{aligned}
\end{equation}
where  $ L_{a} $, $ L_{ide}$, $ L_{dc} $, $ L_{tv}$ represent adversarial loss, identity loss \cite{zhu2017unpaired}, total variation loss and dark channel loss \cite{li2019semi}.

{\bf Student network.} The overall loss functions for the student network are formulated as:
\begin{equation}
\begin{aligned}
L_t = L_{ss} + L_{su}
\end{aligned}
\end{equation}
where $L_{ss}$, $L_{su}$ refers to the loss functions of the supervised and unsupervised branches, respectively.
 \begin{equation}
\begin{aligned}
L_{ss} = \lambda _1*L_{ugs} + \lambda _2*L_a
\end{aligned}
\end{equation}
 \begin{equation}
\begin{aligned}
L_{su} = \lambda _3*L_{ugu} + \lambda _4*L_{dc} + \lambda _5*L_{tv} +\lambda _6*L_{kl}
\end{aligned}
\end{equation}
where $L_{kl}$ denotes the KL divergence loss.

KL loss: Since the intermediate structure of the network tends to extract high-dimensional haze-related features, we choose the 3-th MDB to supply the synthetic hazy image embedding $V_{syn}$ and the real-world image embedding $V_{real}$. And by taking $V_{syn}$ as the pseudo-label of $V_{real}$, these two haze distribution features are transformed into high-dimensional vectors to calculate $L_{kl}$ \cite{cui2022semi}, which can be expressed as:
 \begin{equation}
\begin{aligned}
L_{kl}=KL(Softmax(V_{real}),Softmax(V_{syn}))
\end{aligned}
\end{equation}
by which to enhance the similarity of haze distribution features between synthetic and real-world images.

\section{Experiments}
\label{sec:typestyle}

\subsection{Implementation Details}
\label{ssec:subhead}
Experiments are implemented on Pytorch 1.7 with NVIDIA RTX 3090 GPU and Adam optimizer with parameters $\beta _1 = 0.9 $, $\beta _2 = 0.99 $, $\epsilon =10^{-8}$ to train the network. The teacher network is trained for 100 epochs, in which we update the unsupervised branch once after updating the supervised five times. The student network is trained for 60 epochs, in which we update the supervised branch once after updating the unsupervised five times. In each stage, the learning rate is set to $10^{-4}$ for the first half and then decays linearly to 0 at the end. The batch size is set to 2. The Semi-UFormer is trained with 10,000 paired and 2000 unpaired samples from the OTS and URHI datasets \cite{Benchmarking}. The loss weights are set to: $\lambda _1=1, \lambda _2=10^{-2}, \lambda _3=2, \lambda _4=10^{-2}, \lambda _5=10^{-5}, \lambda _6=10^{-6}$.

\subsection{Comparison Results}
\label{ssec:subhead}

\begin{table}[htbp]
\centering
\footnotesize
\begin{tabular}{*{6}{c}}
  \toprule
  \multirow{2}*{Method} & \multicolumn{2}{c}{SOTS outdoor} & \multicolumn{2}{c}{HSTS}  & \multirow{2}*{Time} \\
  \cmidrule(lr){2-3}\cmidrule(lr){4-5}
  & PSNR↑ & SSIM↑ & PSNR↑ & SSIM↑  \\
  \midrule
  DCP \cite{he2010single} & 18.83 & 0.819 & 17.01 & 0.803 & 1.41\\
  NCP \cite{berman2016non} & 18.07 & 0.802 & 17.62 & 0.798 & 1.45\\
  AOD \cite{li2017aod} & 20.08 & 0.861 & 19.68 & 0.835 & {\bf0.11}\\
  GCA \cite{chen2019gated} & 21.66 & 0.867 & 21.37 & 0.874 & 0.21\\
  EPDN \cite{qu2019enhanced} & 22.57 & 0.863 & 20.37 & 0.877 & 0.23 \\
  GFN-IJCV \cite{zhang2020gated} & 24.21 & 0.849 & 23.17 & 0.829 & 0.43\\
  Semi-dehazing \cite{li2019semi} & 24.79 & 0.892 & 24.36 & 0.889 & 0.32\\
  YOLY \cite{li2021you} & 20.39 & 0.889 & 21.02 & 0.905 & 40.56\\
  PSD \cite{chen2021psd} & 20.49 & 0.844 & 19.37 & 0.824 & 0.39\\
  Ours &  {\bf26.87} &  {\bf 0.928} &  {\bf 28.72} &  {\bf 0.934} & 0.25\\
  \bottomrule
\end{tabular}
	\caption{Averaged PNSR, SSIM, and runtime (CPU/GPU) with state-of-the-art (SOTA) dehazing algorithms on synthetic datasets.}
	\label{tab:1}
\end{table}
{\bf Results on Synthetic Datasets.} Semi-UFormer is evaluated on the SOTS outdoor and HSTS sets\cite{Benchmarking} with nine SOTA dehazing algorithms. As exhibited in Table \ref{tab:1}, our method achieves the highest PSNR and SSIM values on both datasets.

{\bf Results on Real-world Images.} To evaluate the dehazing performance on real-world images, we apply the blind image quality evaluation index SSEQ \cite{liu2014no}, color evaluation index $\sigma$ \cite{Chen_2014_CVPR} and HCC \cite{xu2015review}. Our Semi-UFormer produces the cleanest dehazed images with high visual quality compared with other dehazing approaches, as depicted in Fig. \ref{fig:4} and Table \ref{tab:2}.


\begin{table}[htbp]
\centering
\footnotesize
\begin{tabular}{ c c c c}
   \toprule
       Method     & SSEQ↓ & $\sigma$↓ &  HCC↑  \\
   \midrule
   DCP & 38.402 & 0.0033 & -0.1609 \\
   NCP & 41.506 & 0.0154 & -0.0220 \\
   GCA & 38.400 & 0.0067 & -0.0284\\
   EPDN & 38.632 & \textcolor{blue}{0.0009} & -0.0757\\
   Semi-dehazing & \textcolor{blue}{ 38.104} & 0.0089 & 0.2124\\
   PSD & 40.912 & 0.0017 & \textcolor{red}{0.6085}\\
   Ours & \textcolor{red}{37.751} & \textcolor{red}{0.0001} & \textcolor{blue}{0.2340}\\
   \bottomrule
\end{tabular}
	\caption{ Quantitative comparisons (SSEQ/$\sigma$/HCC) with SOTA approaches on 50 real-world images. \textcolor{red}{ Red } and \textcolor{blue}{blue} colors are used to indicate the $1^{st}$ and $2^{nd}$ ranks, respectively.}
	\label{tab:2}
\end{table}

\begin{table}[htbp]
\centering
\footnotesize
\begin{tabular}{ c c c c c}
   \toprule
       Variants & Base & $V_1$ &  $V_2$ & $V_3$\\
   \midrule
   MDB & w/o & \checkmark & \checkmark & \checkmark \\ 
   Uncertainty & w/o & w/o & \checkmark & \checkmark \\
   KL loss & w/o & w/o & w/o & \checkmark \\
   \midrule
   PSNR & 24.74 & 25.34 & 26.17 & {\bf 26.87}\\ 
   SSIM & 0.905 & 0.910 & 0.919 & {\bf 0.928} \\ 
   \bottomrule
\end{tabular}
	\caption{ Ablation Analysis on Semi-UFormer.}
	\label{tab:3}
\end{table}

\subsection{Ablation Study}
\label{ssec:subhead}
In ablation studies, we first build the base network with the original Dehazeformer-S module trained with $L_1$ loss (replace the uncertainty loss). Then, we build base + MDB module $\rightarrow$ $V_1$, $V_1$ + uncertainty $\rightarrow$ $V_2$, $V_2$ + Knowledge Distillation $\rightarrow$ $V_3$ (full model). As exhibited in Table \ref{tab:3}, our complete network scheme achieves the best dehazing performance.

\section{Conclusion}
\label{sec:majhead}


In this work, a novel semi-supervised uncertainty-aware transformer network called Semi-UFormer is proposed for image dehazing, which leverages both real-world data and uncertainty guidance information to facilitate the dehazing tasks. To bridge the gap between synthetic and real data, we build our Semi-UFormer on top of a knowledge distillation framework and apply a two-branch network to train our model on both synthetic and real-world images. 
Moreover, we exploit an uncertainty estimation block (UEB) to predict the pixel uncertainty of the coarse dehazed results and then guide the network to better restore the image edges and structures. Experiments on both synthetic and real-world images fully validate the effectiveness of our Semi-UFormer.


\bibliographystyle{IEEEbib}
\bibliography{refs}

\end{document}